\documentclass[journal,twoside,web]{ieeecolor}
\usepackage{booktabs}
\usepackage{multirow,graphicx}
\usepackage{tmi}
\usepackage{cite}
\usepackage{amsmath,amssymb,amsfonts}
\usepackage{bm, graphicx, hyperref, mathrsfs,color}
\usepackage{algorithmic}
\usepackage{graphicx}
\usepackage{textcomp}
\def\BibTeX{{\rm B\kern-.05em{\sc i\kern-.025em b}\kern-.08em
    T\kern-.1667em\lower.7ex\hbox{E}\kern-.125emX}}
\begin{document}
\title{Holistic Surgical Phase Recognition\\ with Hierarchical Input Dependent \\State Space Models}
\author{Haoyang Wu, Tsun-Hsuan Wang, Mathias Lechner, Ramin Hasani,  Jennifer A. Eckhoff, Paul Pak,\\ Ozanan R. Meireles, Guy Rosman, Yutong Ban, Daniela Rus
% \thanks{This paragraph of the first footnote will contain the date on which
% you submitted your paper for review. It will also contain support information,
% including sponsor and financial support acknowledgment. For example, 
% ``This work was supported in part by the U.S. Department of Commerce under Grant BS123456.'' }
\thanks{Haoyang Wu and Yutong Ban are from  UM-SJTU Joint Institute, Shanghai Jiao Tong University, Shanghai 200240, China.
(e-mail: william-wu@sjtu.edu.cn, yban@sjtu.edu.cn).}
\thanks{Tsun-Hsuan Wang, Mathias Lechner, Ramin Hasani, Yutong Ban and Daniela Rus are with the Computer Science \& Artificial Intelligence Laboratory, Massachusetts Institute of Technology, Boston, MA 02139, USA.
(e-mail: tsunw@mit.edu, mlechner@mit.edu, rhasani@mit.edu, 
yban@csail.mit.edu, rus@csail.mit.edu).}
\thanks{Mathias Lechner, Ramin Hasani, Paul Pak are with Liquid AI, Boston, MA 02142, USA. (e-mail: mathias@liquid.ai, ramin@liquid.ai, paul@liquid.ai).}
\thanks{Jennifer A. Eckhoff is with University Hospital of Cologne, Cologne 50937,
Germany. (e-mail: jennifer.eckhoff@uk-koeln.de).}
\thanks{Guy Rosman, Ozanan R. Meireles are with the Department of Surgery, Duke University, Durhamm, NC 27707, USA (e-mail: guy.rosman@duke.edu, ozanan.meireles@duke.edu).}
}

\maketitle

\newcommand{\localModels}{LA-SSM}
\newcommand{\globalModels}{GR-SSM}

\begin{abstract}
Surgical workflow analysis is essential in robot-assisted surgeries, yet the long duration of such procedures poses significant challenges for comprehensive video analysis. Recent approaches have predominantly relied on transformer models; however, their quadratic attention mechanism restricts efficient processing of lengthy surgical videos. In this paper, we propose a novel hierarchical input-dependent state space model that leverages the linear scaling property of state space models to enable decision making on full-length videos while capturing both local and global dynamics. Our framework incorporates a temporally consistent visual feature extractor, which appends a state space model head to a visual feature extractor to propagate temporal information. The proposed model consists of two key modules: a local-aggregation state space model block that effectively captures intricate local dynamics, and a global-relation state space model block that models temporal dependencies across the entire video. The model is trained using a hybrid discrete-continuous supervision strategy, where both signals of discrete phase labels and continuous phase progresses are propagated through the network. Experiments have shown that our method outperforms the current state-of-the-art methods by a large margin (+2.8\% on Cholec80, +4.3\% on MICCAI2016, and +12.9\% on Heichole datasets). Code will be publically available after paper acceptance.
\end{abstract}


\section{Introduction}
\IEEEPARstart{S}~urgical phase recognition plays a vital role in computer-assisted surgery (CAS), enhancing both intraoperative decision-making and postoperative analysis \cite{postoperative1}. Accurate phase recognition facilitates the annotation of surgical video workflows and helps identify deviations from established protocols \cite{deviations}. In particular, contemporary advances in deep learning have markedly improved recognition performance \cite{twinanda2016endonetdeeparchitecturerecognition}, underscoring the impact of these methods in surgical phase analysis. Nonetheless, surgical videos present unique challenges: they often extend over several hours and exhibit intricate spatial details at the frame level along with complex temporal dynamics unfolding over both short and long durations. These temporal dynamics can be viewed as comprising two components: local dynamics that capture fine-grained surgical actions, and global dynamics that reflect the overall structure of the surgical phases. Consequently, effectively capturing both the spatial–temporal information and the interplay between these local and global temporal features remains a significant challenge in the field.

\begin{figure}[t!]  
\centering  
\includegraphics[width=\columnwidth]{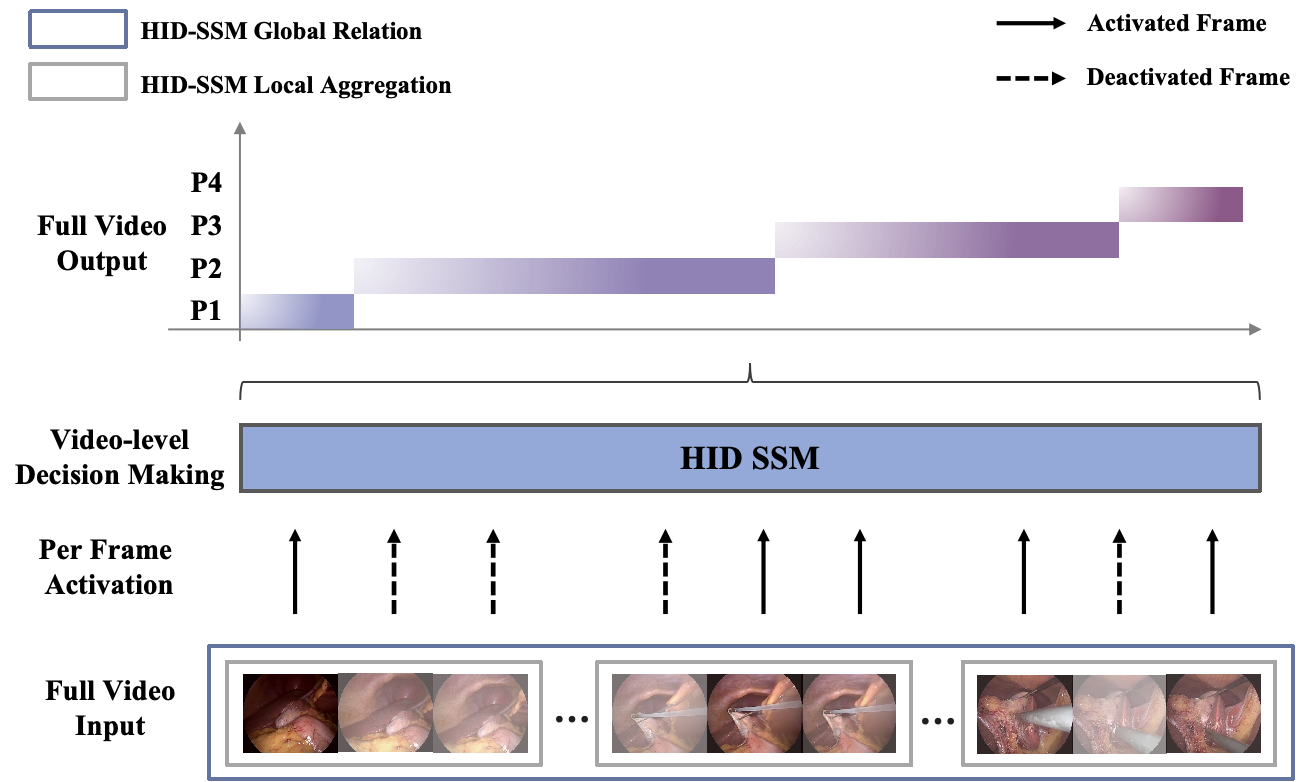} 
\label{teaser}  
\caption{Hierarchical Input-Dependent State Space Model. The proposed method leverages the linear scaling property of State Space Models, allowing the network to process an entire video as input. It selectively activates individual frames to both aggregate intricate local information and capture overarching global context, ultimately generating a full video prediction of the video's phase progression.}
\end{figure} 

Recent studies have investigated the spatial–temporal dynamics of surgical phase recognition by adopting multi-stage paradigms that model spatial and temporal features independently. In these methodologies \cite{jin2021temporalmemoryrelationnetwork,yi2021endtoendexploremultistagearchitecture,gao2021transsvnetaccuratephaserecognition}, a spatial feature extractor is first trained to derive per-frame feature representations, which are subsequently fed into a temporal model to capture inter-frame dynamics. Many of these approaches employed transformer architectures due to their intrinsic capacity to model sequential relationships via attention mechanisms. For instance, \cite{gao2021transsvnetaccuratephaserecognition} utilized self- and cross-attention based transformers to mitigate spatial information loss during temporal modeling. Nevertheless, optimizing transformer-based models for surgical phase recognition remains challenging because of their quadratic computational complexity with respect to sequence length, resulting in substantial memory consumption and impeding scalability for long-duration surgical procedures. To mitigate these constraints, such models often require resource-intensive multi-GPU training~\cite{yang2024surgformersurgicaltransformerhierarchical} or must resort to fixed-size windowing strategies~\cite{gao2021transsvnetaccuratephaserecognition}; however, the former approach imposes excessive computational demands while the latter prevents full-video decision making, thereby restricting the models’ capacity to capture the complete and holistic temporal context present throughout the whole video.

On the other hand, state space models \cite{gu2022efficientlymodelinglongsequences, smith2023simplifiedstatespacelayers, fu2023hungryhungryhipposlanguage} have emerged as a promising alternative for modeling long-range dependencies in sequential data. Unlike transformers, which rely on self-attention, state space models leverage state-space representations to discretely approximate the dynamics of continuous ordinary differential equations, allowing for efficient sequence modeling with linear computational complexity. This advantage makes them particularly suitable for long surgical videos, where comprehensively capturing both local transitions and workflow global temporal structures is essential.

In this paper, we introduce the novel Hierarchical Input-Dependent State Space Model for full-video, online surgical phase recognition. Our approach leverages the state space model to effectively capture spatial-temporal as well as local-global temporal information. Our main contributions can be summarized as follows:  
\begin{itemize}  
    \item We propose an efficient framework capable of holistically capturing information throughout the entire video and generating full-video predictions.
   \item We design a robust hierarchical model that effectively captures both local and global temporal information. By progressively modeling temporal features across different levels, our model ensures precise recognition of fine-grained surgical actions while maintaining an understanding of the overall surgical workflow.
   \item We justify the effectiveness of our model by demonstrating its internal mechanism for selective activation across timesteps, as well as its ability to capture long-term dependencies and retrieve information from hundreds of timesteps in the past.
   \item We conduct extensive experiments on three public surgical datasets, where our model significantly outperforms existing methods, setting the new state-of-the-art performance.
\end{itemize}  

\section{Related Works}
\subsection{Surgical Phase Recognition}
% \textcolor{blue}{Early surgical phase recognition predominantly relied on statistical methods, particularly dynamic Bayesian networks (DBNs) such as Hidden Markov Models (HMMs) \cite{HMM_intro}. Blum et al. \cite{Blum} optimized HMM topologies through model-merging techniques, while Franke et al. \cite{FRANKE2015158} proposed a classification approach leveraging multi-perspective workflow modeling. Despite their effectiveness, these methods rely heavily on detailed annotations of both low-level surgical actions and high-level procedural contexts, making them resource-intensive and application-specific. This limitation has sparkled interest in alternative methodologies. In particular, the advent of deep learning models\cite{Alexnet, girshick2014richfeaturehierarchiesaccurate, long2015fullyconvolutionalnetworkssemantic} have emerged as a promising approach, demonstrating their superior performance across diverse surgical datasets. These models can be characterized into two categories.}
Modern surgical phase recognition primarily relies on deep learning models\cite{twinanda2016endonetdeeparchitecturerecognition,gao2021transsvnetaccuratephaserecognition}, which have demonstrated superior performance across a range of surgical datasets. These models can be broadly categorized into two categories based on their target tasks.
The first category of methods explicitly focuses on the task of surgical phase recognition. \cite{twinanda2016endonetdeeparchitecturerecognition} introduced EndoNet, which employs a fine-tuned AlexNet for spatial feature extraction in surgical workflow analysis. \cite{SV-RCNet} proposed SV-RCNet, integrating ResNet for spatial representation learning and Long Short-Term Memory (LSTM) networks for sequential modeling. TMRNet\cite{jin2021temporalmemoryrelationnetwork} employed a non-local memory bank to incorporate long-term contextual dependencies. \cite{Czempiel_2020} advanced this line with TeCNO, which leveraged causal dilated Multi-stage Temporal Convolution Networks (MS-TCN) \cite{farha2019mstcnmultistagetemporalconvolutional} to enhance the modeling of long-range temporal dependencies. Building upon TeCNO, \cite{gao2021transsvnetaccuratephaserecognition} introduced a three-stage framework that integrated transformer-based \cite{vaswani2023attentionneed} self- and cross-attention mechanisms. This approach retrieved spatial information loss during temporal modeling and improved the model’s ability to capture intricate dependencies across different surgical phases. Our proposed Hierarchical Input-Dependent State Space Model also belongs to this category, further refining phase recognition by incorporating selective state space models to enhance temporal information flow.

Another category of methods employs multitask learning by integrating auxiliary learning objectives to enhance model performance in surgical phase recognition.\cite{twinanda2016endonetdeeparchitecturerecognition} were the first to propose an architecture jointly trained on surgical phase recognition and tool presence detection, leveraging the complementary nature of these tasks to improve phase prediction accuracy.\cite{jin2019multitaskrecurrentconvolutionalnetwork} introduced MTRCNet-CL, which incorporated a correlation loss to explicitly model the relationship between low-level tool feature extraction and high-level phase prediction, facilitating greater interaction between the two tasks.  Aside from tool presence detection, \cite{Ramesh_2021} proposed the Multi-task Multi-Stage Temporal Convolutional Network (MTMS-TCN), which simultaneously predicted high-level surgical phases and low-level procedural steps. By jointly modeling different levels of granularity within surgical workflows, their approach sought to capture both fine-grained step-wise variations and broader phase transitions more effectively.

\subsection{State Space Models}
State Space Models (SSMs) have recently emerged as a powerful architecture for sequence modeling, offering an efficient way to capture long-range dependencies. They achieve this by discretizing an ordinary differential equation using a timescale parameter and encoding long-term information within a state matrix. Based on how the timescale parameter is derived, SSMs can be classified into two categories: Linear Time Invariant systems, where the timescale remains fixed, and Input-Dependent systems, where the timescale dynamically adapts based on the input.
\\
\textbf{Linear Time Invariant (LTI)} Early SSMs \cite{gu2021combiningrecurrentconvolutionalcontinuoustime} typically employed a fixed timescale parameter. As a result, the parameterized state transition matrix remained structured across time steps and was initialized using the HiPPO parameterization \cite{gu2020hipporecurrentmemoryoptimal}. Building on this, \cite{gu2022efficientlymodelinglongsequences} introduced the Structured State Space (S4), allowing each input dimension to evolve independently as a distinct SSM system. To enhance computational efficiency, \cite{smith2023simplifiedstatespacelayers} proposed the Simplified State Space (S5), which improved channel mixing by treating each input as a vector rather than a scalar. \cite{fu2023hungryhungryhipposlanguage} further explored the application of structured SSMs in large language models, significantly narrowing the performance gap with transformers on synthetic language modeling benchmarks. 
\\
\textbf{Input-Dependent SSMs} Although LTI SSMs have demonstrated strong performance, their reliance on a fixed timescale parameter represents a notable limitation. In particular, this mechanism treats all elements in the sequence equally, which restricts the model’s ability to effectively differentiate between irrelevant features and those that are critical. To address this issue and enable varying emphasis across sequences, \cite{hasani2022liquidstructuralstatespacemodels} introduced Liquid-S4. This approach leverages the Liquid Neural Network \cite{hasani2020liquidtimeconstantnetworks}, which is dynamically activated by stimuli to incorporate input-dependent terms, thereby allowing the model to adaptively attend to features across the sequence. Moreover, \cite{gu2024mambalineartimesequencemodeling} proposed the Selective State Space model (S6), known as Mamba. Mamba introduces selectivity mechanisms that enhance information flow over long sequences, striking a balance between short-term precision and long-term coherence. Building on this, \cite{dao2024transformersssmsgeneralizedmodels} introduced the State Space Duality framework (Mamba-2), establishing theoretical and practical connections between SSMs and transformers, thereby unifying the strengths of both architectures. While Mamba and Mamba-2 are inherently causal models, limiting their ability to leverage future context, \cite{hwang2024hydrabidirectionalstatespace} addressed this limitation by introducing Hydra, which extended Mamba-2’s matrix mixer framework to enable bidirectional processing, allowing for intuitive contextual reasoning. Some early explorations of applying SSMs in surgical phase recognition were conducted in \cite{cao2024srmambaeffectivesurgicalphase}, which demonstrates the effectiveness of SSMs in understanding the surgical phases. However, their approach has yet to fully harness the capabilities of SSMs.

% \textcolor{blue}{To fully leverage the potential of SSMs in modeling long sequences, early attempts have been made to apply SSMs in surgical phase recognition. Cao et al. \cite{cao2024srmambaeffectivesurgicalphase} proposed SR-Mamba, which demonstrates the effectiveness of SSMs in making video-level decisions. However, their approach has yet to fully harness the capabilities of SSMs. Specifically, the method directly feeds features extracted by the spatial feature extractor into the SSM for comprehensive modeling, without refining the local and global dynamics of these features. Additionally, the implementation of a basic bidirectional Mamba model does not effectively integrate contextual information to the extent achieved by Hydra. Furthermore, the potential of SSMs in causal surgical video modeling remains underexplored, despite its significant real-world applications.}
\section{Background}
% Introduce SSMs mathematically
State Space Models are defined by a continuous system that map an input signal \( u \in \mathbb{R}^L \) to a hidden state \( x \in \mathbb{R}^{L \times N} \), which is subsequently projected onto the output \( y \in \mathbb{R}^L \), $L$ represents the sequence length and N represents the state dimension. The system of equations is expressed as:
\begin{equation}
    x'(t)= \textbf{A}x(t-1) + \textbf{B}u(t),\quad\quad y(t)= \textbf{C}x(t)
\label{SSM_original}
\end{equation}
\( \textbf{A} \in \mathbb{R}^{N \times N} \) is the state matrix, \( \textbf{B} \in \mathbb{R}^{N} \) is the input matrix, and \( \textbf{C} \in \mathbb{R}^{N} \) is the output matrix.
\\\textbf{Discretization} The discretization process transforms SSM's continuous system into its discrete counterpart. Several discretization methods, such as bilinear transformation and zero-order hold (ZOH), can be employed to achieve this. These methods define the discrete state matrix \( \overline{\textbf{A}} = f_A (\textbf{A}, \Delta) \) and input matrix \( \overline{\textbf{B}} = f_B (\textbf{A}, \textbf{B}, \Delta) \), where \( f_A \) and \( f_B \) represent the respective discretization rules applied to the continuous-time state and input matrices. The resulting discrete-time equations can be expressed as:
\begin{equation}
    x_t= \overline{\mathbf{A}}x_{t-1} + \overline{\mathbf{B}}u_t,\quad\quad y_t= \mathbf{C}x_t
\label{SSM_general}
\end{equation}
\\\textbf{Recurrent Computation} After discretizing the SSM system, we can further unroll \eqref{SSM_general}, and express the output as:
\setcounter{equation}{0}  % Reset the equation counter
\renewcommand{\theequation}{3.\alph{equation}}  % Modify the equation numbering to 3.a, 3.b, 3.c
\begin{align}
y_0 &= \mathbf{C}\overline{\mathbf{B}} \, u_0, \label{eq:y0} \\
y_1 &= \mathbf{C}\overline{\mathbf{A}\mathbf{B}} \, u_0 + \mathbf{C}\overline{\mathbf{B}} \, u_1, \label{eq:y1} \\
y_2 &= \mathbf{C}\overline{\mathbf{A}}^2 \overline{\mathbf{B}} \, u_0 + \mathbf{C}\overline{\mathbf{A}\mathbf{B}} \, u_1 + \mathbf{C}\overline{\mathbf{B}} \, u_2. \label{eq:y2}
\end{align}
To accelerate the recurrent computation, \cite{gu2024mambalineartimesequencemodeling} proposed a hardware-aware algorithm that optimizes the utilization of modern GPUs' memory and computational resources. Notably, the computation of SSMs can also be expressed in a more general matrix mixer form \cite{dao2024transformersssmsgeneralizedmodels}, which is detailed in the next section.
%\\\textbf{Convolutional Representation} \textcolor{blue}{Although \eqref{SSM_general} itself constitutes a recurrent representation of the state-space model (SSM). By further unrolling \eqref{SSM_general}, the output can be explicitly expressed as:}
%\setcounter{equation}{0}  % Reset the equation counter
%\renewcommand{\theequation}{3.\alph{equation}}  % Modify the equation numbering to 3.a, 3.b, 3.c
%\begin{align}
%y_0 &= \mathbf{C}\overline{\mathbf{B}} \, u_0, \label{eq:y0} \\
%y_1 &= \mathbf{C}\overline{\mathbf{A}\mathbf{B}} \, u_0 + \mathbf{C}\overline{\mathbf{B}} \, u_1, \label{eq:y1} \\
%y_2 &= \mathbf{C}\overline{\mathbf{A}}^2 \overline{\mathbf{B}} \, u_0 + \mathbf{C}\overline{\mathbf{A}\mathbf{B}} \, u_1 + \mathbf{C}\overline{\mathbf{B}} \, u_2. \label{eq:y2}
%\end{align}\textcolor{blue}{As noted by Gu et al.~\cite{gu2024mambalineartimesequencemodeling}, this formulation can be equivalently expressed in a vectorized form as a convolutional representation. Specifically, by expanding across sequence length, the convolutional kernel is given by:}
%\setcounter{equation}{3}
%\renewcommand{\theequation}{\arabic{equation}}
%\begin{equation}
%    \mathbf{K} = (\mathbf{C}\overline{\mathbf{B}}, \mathbf{C}\overline{\mathbf{AB}}, \mathbf{C}\overline{\mathbf{A}}^2 \overline{\mathbf{B}}, \dots).
%    \label{eq:K_kernel}
%\end{equation}
%\textcolor{blue}{By performing elementwise multiplication between $\mathbf{K}$ and the input sequence $u$, the output sequence can be efficiently computed.}

\begin{figure*}[ht!]
\centering
\includegraphics[width=\textwidth]{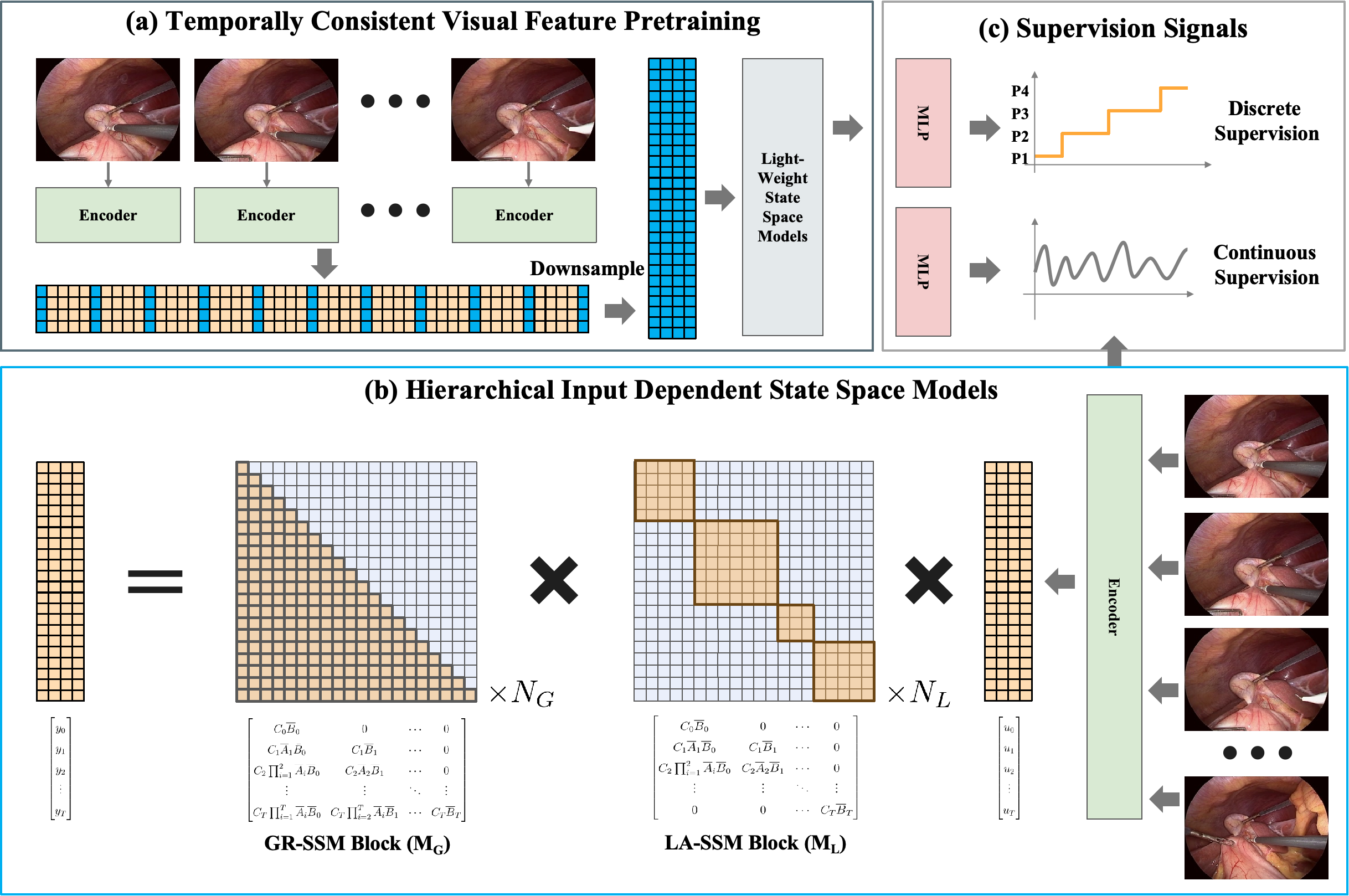}
\caption{Hierarchical Input Dependent State Space Model Overview.(a) A temporally consistent visual feature extractor that incorporates temporal information in feature extraction by learning the structure of the downsampled surgical videos; (b) The hierarchical input-dependent state-space model that takes the extracted features as input and enhances both local and comprehensive temporal relations; (c) The supervision signals that are utilized to train the model, comprising both a discrete and a continuous signal.}
\label{fig1}
\end{figure*}

\section{Methods}
In this section, we present the details of the proposed method. We start from the problem formulation, then the temporally consistent feature extractor, the hierarchical input-dependent state-space model, the discrete-continuous supervision, and finally the temporal selective activation.

\subsection{Problem Formulation}
We formulate surgical phase recognition as an end-to-end long video understanding problem. Current approaches to recognize surgical phases predominantly rely on sequential input frames extracted from surgical video streams at a reduced temporal resolution. Since surgical videos can span several hours, training models in an end-to-end manner presents substantial computational and optimization challenges. Consequently, the most widely adopted strategy follows a two-stage framework, wherein a spatial feature extractor and a temporal model are trained independently, with gradient flow explicitly truncated between them.

%Current approaches to surgical phase recognition predominantly rely on sequential input frames extracted from surgical video streams at a reduced temporal resolution. Since surgical videos can span several hours, training models in an end-to-end manner presents substantial computational and optimization challenges. Consequently, the most widely adopted strategy follows a two-stage framework, wherein a spatial feature extractor and a temporal model are trained independently, with gradient flow explicitly truncated between them.

Let \(\mathbf{v} \in \mathbb{R}^{T \times C \times H \times W}\) represents a uniformly down-sampled surgical video frames, where \(T\) denotes the sequence length, \(C\) is the number of channels, and \(H\) and \(W\) are the spatial dimensions. The two-stage framework first processes \(\mathbf{v}\) through a feature extractor \(\mathcal{F}\), producing a sequence of feature vectors \(\mathbf{u} \in \mathbb{R}^{T \times D}\), where \(D\) denotes the feature dimension. These feature representations are then passed through a temporal model \(\mathcal{T}\), yielding the final phase predictions \(\mathbf{\hat{z}} \in \mathbb{R}^{T \times N_p}\), where \(N_p\) represents the number of surgical phases. The predicted phase at each timestep is obtained via an \(\textit{argmax}\) operation and compared against the ground truth labels \(\mathbf{z}\in R^T\). Thus, the two-stage surgical phase recognition pipeline can be formally expressed as:

\begin{equation}
    \mathbf{u} = \mathcal{F}(\mathbf{v}) \in \mathbb{R}^{T \times D}, \quad \mathbf{\hat{z}} = \mathcal{T}(\mathbf{u}) \in \mathbb{R}^{T \times N_p}.
\label{eq:formulation}
\end{equation}

\subsection{Temporally Consistent Visual Feature Extraction}
Most of the feature extractors in prior works \cite{SV-RCNet,gao2021transsvnetaccuratephaserecognition} are based on per-frame static visual backbones, and are pre-trained in a per-frame fashion. Therefore, the \textit{inter-frame temporal dependencies are missing}. We claim that for surgical long-video understanding tasks, incorporating temporal information at the feature extraction stage is equally essential, since it enables the spatial model to focus on salient spatial features that contribute to a coherent temporal representation.

% In prior works, most feature extractors \cite{SV-RCNet,gao2021transsvnetaccuratephaserecognition} are based on Convolutional Neural Networks (CNNs) \cite{CNN}, such as ResNet \cite{he2015deepresiduallearningimage}, which operate on a per-frame basis, thereby disregarding inter-frame temporal dependencies. We claim that incorporating temporal information at the feature extraction stage is equally essential, as it enables the spatial model to focus on salient spatial features that contribute to a coherent temporal representation.

To address this limitation, we design a spatial-temporal pre-training stage. We use a Swin Transformer \cite{liu2021swintransformerhierarchicalvision} as the visual backbone. We then integrate it with a light-weight temporal model, in our case, a shallow state-space model (SSM) with \( N_s \) layers. The output of the temporal model is further fed into a fully-connected layer to predict the phase labels, which follows the traditional paradigm of a spatial-temporal video understanding framework.

We perform a temporal sparsification process to enhance the feature representation with richer temporal information while keeping a reasonable memory cost. To be specific, we first reduce the number of frames by selecting them at fixed intervals (e.g., every 5th frame). This interval-based selection is then combined with a sliding window that shifts the starting frame for each training iteration. To note that, the model is pre-trained for only one epoch, to have an efficient training pipeline. This paradigm allows for effectively and efficiently propagating temporal information back into the visual backbone. After the pre-training, we drop the temporal model (e.g., the shallow state space model), since it only models the high-level temporal dependencies roughly. For future feature extractions, we only keep the visual backbone, which is expected to contain the temporally consistent visual features. The granular temporal dependencies between the frames are modeled by the hierarchical input-dependent state space models in the next section.

\subsection{Hierarchical Input-dependent State Space Model}
\label{HID-SSM}
Although the rough temporal information is captured in extracted features, the fine-grained inter-frame dependencies are still missing. In this section, we introduce a Hierarchical Input Dependent State Space Model (HID-SSM) that \textit{operates at multiple temporal scales to effectively retrieve the fine-grained local temporal dynamics like the surgical actions, and consolidate the global temporal information in surgery videos}.

\noindent \textbf{State Space Model Block} The proposed HID-SSM model is built upon different state space model blocks, the Local Aggregation (LA) blocks for local surgical action extraction, and the Global Relation (GR) blocks for long-term dependency modeling. We use  $\mathcal{SS}$ to represent each SSM block. Within each SSM block, we have a causal version for one-directional information propagation and a contextual version for the bidirectional setup. The detailed formulation of each state space model block is shown as follows.

 To model the continuous surgical videos with input-dependent state space models, we need to perform a discretization process. We dynamically parameterize the timescale \(\Delta\) as a function of the current surgical frame, thereby allowing per frame activation across timesteps. To guarantee stable and efficient state transitions, the zero-order hold method is applied to each block for discretization. In particular, the timescale and matrix coefficients \cite{gu2024mambalineartimesequencemodeling} in \eqref{SSM_general} are computed as:
 % The proposed HID-SSM model is built upon different state space model blocks (See \textcolor{red}{Fig.}), denoted as $\mathcal{SS}$. Specifically, within each ID-SSM block, we utilize Mamba-2 \cite{dao2024transformersssmsgeneralizedmodels} for causal HID-SSM and Hydra \cite{hwang2024hydrabidirectionalstatespace} for contextual HID-SSM. Both Mamba-2 and Hydra dynamically parameterize the timescale \(\Delta\) as a function of the current input, thereby dynamically takes in stimuli across timesteps. To guarantee stable and efficient state transitions, the models use the Zero-Order Hold (ZOH) discretization method. In particular, the timescale and matrix coefficients noted in \eqref{SSM_general} are computed as:
 \setcounter{equation}{0} % Reset counter to 0
\renewcommand\theequation{6.\alph{equation}}
\begin{align}
    \Delta_t &= \text{softplus}(\text{Broadcast}(S_{\Delta})) \label{eq:6.a-Dt} \\
    \overline{\mathbf{A}}_t &= \exp(\Delta_t S_A) \label{eq:6.b-DA} \\
    \overline{\mathbf{B}}_t &= (\Delta_t S_A)^{-1}\Bigl(\exp(\Delta_t S_A) - I\Bigr) \cdot \Delta_t S_B \label{eq:6.c-DB} \\
    \mathbf{C}_t &= S_C \label{eq:6.d-DC}
\end{align}
 \setcounter{equation}{6}
\renewcommand\theequation{\arabic{equation}}

Here, \(S_{(\cdot)}\) denotes the initialization for the parameter $(\cdot)$. For $S_A$, we use a simple scalar time identity initialization \cite{gu2022parameterizationinitializationdiagonalstate}, \(S_A = aI\). The other parameters (\(S_{\Delta}\ ,S_B\ ,S_C\)) are initialized from the current surgical frame input \(u_t\) as:
\begin{align}
    S_{(\cdot)} &= \text{Linear}_{(\cdot)}(u_t)
\end{align}
where \(\text{Linear}_{(\cdot)}\) denotes a linear projection into a corresponding space. Therefore, rolling out \eqref{SSM_general} gives:
\begin{subequations}\label{eq:8}
\setcounter{equation}{0} % Reset counter to 0
  \renewcommand\theequation{8.\alph{equation}}
  \begin{align}
    y_0 &= C_0\overline{B}_0u_0,     \label{eq:8a}\\
    y_1 &= C_1\overline{A}_1\overline{B}_0u_0 + C_1\overline{B}_1u_1, 
                                        \label{eq:8b}\\
    &\vdots\nonumber\\
    y_t &= C_t\sum_{i=0}^t \prod_{j=i+1}^{t}\overline{A}_j\overline{B}_iu_i.
                                        \label{eq:8c}
  \end{align}
\end{subequations}
\setcounter{equation}{8}
\renewcommand\theequation{\arabic{equation}}
According to this equation, each state space model block models the relation between the input feature sequence $\mathbf{u}$ and the output sequence $\mathbf{y}$. Based on such a relation, we design a global SSM block and a local-aggregation SSM block for comprehensive temporal dependencies and short-term action dynamics, detailed in the next section. 

\noindent \textbf{Global-Relation SSM Block (\globalModels):}  
We now consider a higher-level structure that operates on the entire surgical video sequence and outputs a temporally correlated feature. The GR-SSM is composed of \(N_G\) sequential SSM blocks, where \(N_G\) is a hyperparameter. It takes as input the whole surgical video feature \(\mathbf{u_{0:T}}\), further derives the output sequence with \eqref{eq:8}, which is rewritten into the matrix mixer form \cite{dao2024transformersssmsgeneralizedmodels} \refstepcounter{equation}\label{eq9:GR-SSM}
$\mathbf{y} = \mathbf{M_G\times u}$~(\theequation):
\begin{equation*}
\resizebox{\columnwidth}{!}{$
\renewcommand{\arraystretch}{1.5}
\left[\begin{array}{@{}c@{}}
y_0 \\
y_1 \\
y_2 \\
\vdots \\
y_T
\end{array}\right]
=
\left[\begin{array}{@{}cccc@{}}
C_0\overline{B}_0 & 0  & \cdots & 0 \\
C_1\overline{A}_1\bar{B}_0 & C_1\overline{B}_1 & \cdots & 0 \\
C_2\prod_{i=1}^{2} \overline{A}_i \overline{B}_0 & C_2\overline{A}_2\overline{B}_1 & \cdots & 0 \\
\vdots & \vdots & \ddots & \vdots \\
C_T\prod_{i=1}^T\overline{A}_i\overline{B}_0 & C_T\prod_{i=2}^T\overline{A}_i\overline{B}_1 & \cdots & C_T\overline{B}_T
\end{array}\right]
\left[\begin{array}{@{}c@{}}
u_0 \\
u_1 \\
u_2 \\
\vdots \\
u_T
\end{array}\right]
\renewcommand{\arraystretch}{1}
$}
\end{equation*}
where \(\mathbf{M_G}\) denotes the matrix mixer for \globalModels. After computing input feature \(\mathbf{u}\), we further enhance comprehensive temporal modeling through employing a \globalModels. The model is composed of \(N_G\) sequential SSM blocks, where \(N_G\) is a hyperparameter. 

\noindent \textbf{Local-Aggregation SSM Block (\localModels):} 
While the GR-SSM models the global temporal dynamics, it still requires a model to account for local dependencies in neighboring frames. To address this limitation, we introduce a local aggregation state space model block for short-term action recognition. Similar to the GR-SSM, each LA-SSM layer can be expressed in a block-diagonal matrix mixer form \refstepcounter{equation}\label{eq10:LA-SSM}
$\mathbf{y}=\mathbf{M_L \times u}$~(\theequation):
\begin{equation*}
\resizebox{\columnwidth}{!}{$
\renewcommand{\arraystretch}{1.5}
\left[\begin{array}{@{}c@{}}
y_0 \\
y_1 \\
y_2 \\
\vdots \\
y_T
\end{array}\right]
=
\left[\begin{array}{@{}cccc@{}}
C_0\overline{B}_0 & 0 & \cdots & 0 \\
C_1\overline{A}_1\overline{B}_0 & C_1\overline{B}_1 & \cdots & 0 \\
C_2\prod_{i=1}^{2} \overline{A}_i \overline{B}_0 & C_2\overline{A}_2\overline{B}_1 & \cdots & 0 \\
\vdots & \vdots & \ddots & \vdots \\
0 & 0 & \cdots & C_T\overline{B}_T
\end{array}\right]
\left[\begin{array}{@{}c@{}}
u_0 \\
u_1 \\
u_2 \\
\vdots \\
u_T
\end{array}\right]
\renewcommand{\arraystretch}{1}
$}
\end{equation*}

The difference between the matrix mixer \(\mathbf{M_L}\) and the matrix mixer \(\mathbf{M_G}\) is that the \(\mathbf{M_L}\) is in a block diagonal form, as shown in Fig.~\ref{fig1}. For \(\mathbf{M_L}\), each of the \(N_p\) diagonal blocks represents an independent ID-SSM module, where each block is responsible for processing a time segment of the input sequence. 
To be specific, within a \localModels\ layer, each input segment \(\mathbf{u_{t:t+i}}\) is processed by an independent SSM block to extract local temporal dynamics within that time segment. After \(N_L\) stacked \localModels\ layers, the outputs of all blocks are concatenated and passed through a feed-forward network to fuse the segmented representations, yielding the final refined feature output.

% To be specific, let \(\mathbf{u}\) denote the full feature sequence, and let \(\mathbf{u}_{i=1}^{N_p}\) represent its segmentation into \(N_p\) pseudo-phases. Inside a \localModels\ layer, each segment \(\mathbf{u}_i\) is processed by an independent ID-SSM block \(\mathcal{SS}^i\) to extract local temporal dynamics within that pseudo-phase. After applying \(N_L\) stacked \localModels\ layers, the outputs of all blocks are concatenated and passed through a feed-forward network (FFN) to mix the segmented representations, yielding the final refined feature output.

\noindent \textbf{Phase Proposal Network (PPN):} To determine the window size for each time segment in LA-SSM, we employ a phase proposal network to segment the input sequence into semantically coherent pseudo-phases. Such a proposal ensures that each SSM block of the LA-SSM layer receive a visually consistent temporal structure, facilitating the meaningful aggregation of local features. Specifically, the PPN first models the coarse temporal dynamics of the input feature sequence \(\mathbf{u}\) by applying a stack of \(N_q\) ID-SSM layers, yielding an intermediate representation of the same dimension as \(\mathbf{u}\). This representation is subsequently passed through an MLP layer to produce a pseudo-phase prediction \(\hat{\mathbf{z}}_p \in \mathbb{R}^{T \times N_p}\). The PPN is trained in a supervised manner using the cross-entropy loss between the predicted phase labels \(\hat{\mathbf{z}}_p\) and the ground truth annotations \(\mathbf{z}\). Simultaneous to the training of PPN, the obtained pseudo-phases are used to determine the window size of LA-SSM. The window size is set with the length of each pseudo-phase, to ensure that each \localModels\ block \(\mathcal{SS}^i\) processes video segments that are locally consistent.

% Simultaneous to the training of PPN, its output \(\hat{\mathbf{z}}_p\) is used to segment the input video. To be specific, we begin by counting the number of predictions for each phase in \(\hat{\mathbf{z}}_p\), resulting in a rough phase distribution \(\{O_i\}_{i=1}^{N_p}\). Since empirically, the surgical phases are arranged in a sequential order, we partition the input sequence accordingly: the first \(O_1\) frames are assigned to \(\mathcal{SS}^1\), the next \(O_2\) frames to \(\mathcal{SS}^2\), and so on. This ensures that each \localModels\ block \(\mathcal{SS}^i\) processes video segments that are locally consistent, which facilitates its performance.

%\textcolor{red}{To estimate the proportion of each phase within the video, we define:}
%\[
%h_k = \sum_{t=1}^{T} \delta\big(\operatorname{argmax}(\mathbf{P_R}^{t}), k\big), \quad \forall k \in \{0, 1, \dots, K-1\},
%\]
%where the \(\delta\) denotes the Kronecker delta function.

%Assuming that the phases occur sequentially, we segment the video into pseudo-phases based on the cumulative counts \(h_k\). Specifically, the \(k\)-th pseudo-phase segment is given by:
%\[
%\mathbf{u}^k = \mathbf{u}\Bigl[\sum_{j=0}^{k-1} h_j : \sum_{j=0}^{k} h_j \Bigr], \quad \forall k \in \{0, 1, \dots, K-1\}.
%\]
%This segmentation ensures that each pseudo-phase \(V^k\) contains a temporally coherent subset of frames corresponding to the predicted phase \(k\).

\begin{table*}[htbp]
    \centering
    \caption{Results comparison with different state-of-the-art methods on Cholec80 Dataset}
    \resizebox{.8\textwidth}{!}{
    \begin{tabular}{l c c c c c c}
        \toprule
         \textbf{Method} & \textbf{Causality} & \textbf{Architechture} & \textbf{Accuracy $\uparrow$} & \textbf{Precision $\uparrow$} & \textbf{Recall $\uparrow$} & \textbf{Jaccard $\uparrow$} \\
        \midrule
        PhaseNet \cite{twinanda2016miccai} & Causal & CNN & 78.8 $\pm$ 4.7 & 71.3 $\pm$ 15.6 & 76.6 $\pm$ 16.6 & - \\
        EndoNet \cite{twinanda2016endonetdeeparchitecturerecognition} & Causal & CNN & 81.7 $\pm$ 4.2 & 73.7 $\pm$ 16.1 & 79.6 $\pm$ 7.9 & - \\
        SV-RCNet \cite{SV-RCNet} & Causal &  CNN+LSTM & 85.3 $\pm$ 7.3 & 80.7 $\pm$ 7.0 & 83.5 $\pm$ 7.5 & - \\
        OHFM \cite{OHFM} & Causal & CNN+LSTM & 87.3 $\pm$ 5.7 & - & - & 67.0 $\pm$ 13.3 \\
        MTRCNet-CL \cite{jin2019multitaskrecurrentconvolutionalnetwork} & Causal & CNN+LSTM & 89.2 $\pm$ 7.6 & 86.9 $\pm$ 4.3 & 88.0 $\pm$ 6.9 & - \\
        TMRNet \cite{jin2021temporalmemoryrelationnetwork} & Causal & CNN+LSTM & 90.1 $\pm$ 7.6 & 90.3 $\pm$ 3.3 & 89.5 $\pm$ 5.0 & 79.1 $\pm$ 5.7 \\
        TeCNO \cite{Czempiel_2020} & Causal & CNN+TCN & 88.6 $\pm$ 7.8 & 86.5 $\pm$ 7.0 & 87.6 $\pm$ 6.7 & 75.1 $\pm$ 6.9 \\
        Trans-SVNet \cite{gao2021transsvnetaccuratephaserecognition} & Causal & Transformer & 90.3 $\pm$ 7.1 & 90.7 $\pm$ 5.0 & 88.8 $\pm$ 7.4 & 79.3 $\pm$ 6.6 \\
        LoViT \cite{liu2023lovitlongvideotransformer} & Causal & Transformer & 92.4 $\pm$ 6.3 & 89.9 $\pm$ 6.1 & 90.6 $\pm$ 4.4 & 81.2 $\pm$ 9.1 \\
        LAST
        SKiT \cite{SKiT}& Causal & Transformer & 93.4 $\pm$ 5.2 & 90.9 & 91.8& 82.6 \\  
        Surgformer \cite{yang2024surgformersurgicaltransformerhierarchical} & Causal & Transformer & 93.4 $\pm$ 6.4 & 91.9 $\pm$ 4.7 & 92.1 $\pm$ 5.8 & 84.1 $\pm$ 8.0 \\
        S5 \cite{smith2023simplifiedstatespacelayers} & Causal & SSM &87.5 $\pm$  7.3 & 84.8 $\pm$ 5.3 & 84.8 $\pm$ 9.2 & 71.3 $\pm$ 9.1 \\
        Mamba \cite{gu2024mambalineartimesequencemodeling} & Causal & SSM &87.5 $\pm$  7.6 & 84.6 $\pm$ 4.9 & 84.1 $\pm$ 9.8 & 70.9 $\pm$ 10.4 \\
        Mamba-2 \cite{dao2024transformersssmsgeneralizedmodels} & Causal & SSM &87.9 $\pm$  7.5 & 85.4 $\pm$  6.0 & 85.1 $\pm$ 10.1 & 71.8 $\pm$ 9.5 \\
        SR-Mamba \cite{cao2024srmambaeffectivesurgicalphase} & Contextual & SSM & 92.6 $\pm$ 8.6 & 90.3 $\pm$ 5.2 & 90.6 $\pm$ 7.2 & 81.5 $\pm$ 8.6\\ \cmidrule(lr){1-7}
        HID-SSM (Ours) & Causal & SSM & 94.5 $\pm$ 4.8 & 92.1 $\pm$ 5.0 & 91.1 $\pm$ 8.4 & 83.8 $\pm$ 10.0 \\
        HID-SSM (Ours) & Contextual & SSM & \textbf{96.2 $\pm$ 2.9} & \textbf{93.3 $\pm$ 5.5} & \textbf{93.1 $\pm$ 6.1} & \textbf{86.5 $\pm$ 8.7} \\
        \bottomrule
    \end{tabular}
    }
    \label{tab:cholec80}
\end{table*}

\subsection{Discrete-Continuous Supervision}  
Recognizing that SSMs are in fact continuous Neural ODE models, we therefore train the HID-SSM with both discrete and continuous supervision. Specifically, both a discrete loss and a continuous loss are propagated through the network. Given the final encoded feature produced by the \globalModels, this feature is passed into two separate MLP heads: one for discrete classification labels and one for continuous phase progress. The classification head maps the output of the \globalModels\ to a discrete phase prediction \(\mathbf{\hat{z}}_{\text{cls}} \in \mathbb{R}^{T \times N_p}\). In addition, the phase progress head outputs a progress estimate \(\hat{\mathbf{z}}_{\text{prs}} \in (0, 1)\) within the corresponding phase segment. This is compared against the ground truth \(\mathbf{z}_{\text{prs}}\), which is computed by linearly mapping each frame's position within its phase segment to a normalized value between 0 and 1. The total loss function \(\mathcal{L}_{\text{total}}\) is a weighted combination of the discrete and continuous losses:
\begin{equation}
    \mathcal{L}_{\text{total}} = \alpha \mathcal{L}_{\text{CE}}(\mathbf{\hat{z}}_{\text{cls}}, \mathbf{z}_{\text{cls}}) + (1 - \alpha) \mathcal{L}_{\text{MSE}}(\mathbf{\hat{z}}_{\text{prs}}, \mathbf{z}_{\text{prs}})
    \label{eq11:total_loss}
\end{equation}
where $\mathcal{L}_{\text{CE}}$ is the cross entropy loss, $\mathcal{L}_{\text{MSE}}$ is the MSE loss, and  \(\alpha\) is a hyperparameter that controls the relative importance of the two losses. By jointly optimizing these two supervision signals, HID-SSM is able to effectively learn both discrete phase transitions and continuous progress tracking, enabling more accurate temporal modeling of video sequences.

\subsection{Temporal Selective Activation}
For surgical phase recognition, interpretability is a key property of an effective model. Ideally, the model should highlight which specific frames contribute to its decisions. ID-SSM achieves this through temporally selective activation of features. As observed by \cite{gu2024mambalineartimesequencemodeling}, ID-SSMs exhibit a crucial per-frame activation property, which is linked to an input-dependent selection mechanism, introduced through the timescale parameter $\Delta_t$. Since $S_A$ is always initialized negatively, we consider the case where $N = 1$, $S_A = -1$, and $S_B = 1$. Under these conditions, the discretized forms of $\overline{A}_t$ and $\overline{B}_t$ in \eqref{eq:6.b-DA} \eqref{eq:6.c-DB} can be expressed as:

\setcounter{equation}{0} % Reset counter to 0
\renewcommand\theequation{12.\alph{equation}}
\begin{align}  
    \overline{\mathbf{A}}_t &= \exp(-\Delta_t) \label{eq12a:dtA} \\
    \overline{\mathbf{B}}_t &= 1 - \exp(-\Delta_t) \label{eq12b:dtB}
\end{align}  
\setcounter{equation}{12}
\renewcommand\theequation{\arabic{equation}}

This leads to a simplified form of the general state-space model, given by:

\begin{equation}
    x_t = \exp(-\Delta_t) x_{t-1} + (1 - \exp(-\Delta_t)) u_t.
    \label{eq:dt}
\end{equation}

The resulting expression highlights an important property of the model: as $\Delta_t$ increases, the coefficient associated with the previous state $x_{t-1}$ decreases, while the coefficient on the current input $u_t$ increases. This suggests that, with larger $\Delta_t$, the model effectively "resets" the current state, giving greater attention to the current input. Conversely, when $\Delta_t$ is smaller, the model tends to preserve the current state and neglect the current input.

\section{Experiments}

\begin{table*}[htbp]
    \centering
    \caption{Relaxed results of different state-of-the-art methods on MICCAI2016 Dataset}
    \resizebox{.8\textwidth}{!}{
    \begin{tabular}{l c c c c c c}
        \toprule
        \small
        \textbf{Method} & \textbf{Causality} & \textbf{Pipeline} & \textbf{Accuracy $\uparrow$} & \textbf{Precision $\uparrow$} & \textbf{Recall $\uparrow$} & \textbf{Jaccard $\uparrow$} \\
        \midrule
        PhaseNet \cite{twinanda2016miccai} & Causal & CNN & 79.5 $\pm$ 12.1 & - & - & 64.1 $\pm$ 10.3 \\
        SV-RCNet \cite{SV-RCNet} & Causal & CNN+LSTM & 81.7 $\pm$ 8.1 & 81.0 $\pm$ 8.3 & 81.6 $\pm$ 7.2 & 65.4 $\pm$ 8.9 \\
        OHFM \cite{OHFM} & Causal & CNN+LSTM & 85.2 $\pm$ 7.5 & - & - & 68.8 $\pm$ 10.5 \\
        TeCNO \cite{Czempiel_2020} & Causal & CNN+TCN & 86.1 $\pm$ 10.0 & 85.7 $\pm$ 7.0 & 88.9 $\pm$ 4.5 & 74.4 $\pm$ 7.2 \\
        Trans-SVNet \cite{gao2021transsvnetaccuratephaserecognition}& Causal & Transformer & 87.2 $\pm$ 9.3 & 88.0 $\pm$ 6.7 & 87.5 $\pm$ 5.5 & 74.7 $\pm$ 7.7 \\
        S5 \cite{smith2023simplifiedstatespacelayers} & Causal & SSM & 83.9 $\pm$ 9.6 & 82.8 $\pm$ 8.6 & 83.9 $\pm$ 7.5 & 68.9 $\pm$ 8.0 \\
        Mamba
        \cite{gu2024mambalineartimesequencemodeling} & Causal & SSM &82.8 $\pm$  9.8 & 81.6 $\pm$  10.4 & 82.7 $\pm$  7.4 & 67.4 $\pm$  9.0 \\
        Mamba-2 \cite{dao2024transformersssmsgeneralizedmodels} & Causal & SSM &82.6 $\pm$  9.7 & 81.3 $\pm$  10.3 & 82.2 $\pm$  7.5 & 66.8 $\pm$  8.6 \\
        \midrule
        HID-SSM (Ours) & Causal & SSM & \textbf{91.5 $\pm$ 9.0} & 90.2 $\pm$ 6.0 & \textbf{89.8 $\pm$ 7.1} & 80.2 $\pm$ 10.3 \\
        HID-SSM (Ours) & Contextual & SSM & 91.1 $\pm$ 8.8 & \textbf{90.7 $\pm$ 6.6}  & 89.1 $\pm$ 7.8 & \textbf{81.7 $\pm$ 8.5} \\
        \bottomrule
    \end{tabular}
    }
    \label{tab:m2cai}
    \vspace{-4ex}
\end{table*}

\begin{table}[htbp]
    \centering
    \caption{Unrelaxed results of different state-of-the-art methods on Heichole Dataset}
    \resizebox{\columnwidth}{!}{
    \begin{tabular}{l c c c}
        \toprule
        \textbf{Method} & \textbf{Causality} & \textbf{Pipeline} & \textbf{F1 Score $\uparrow$}\\
        \midrule
        TeCNO \cite{Czempiel_2020} & Causal & CNN+TCN & 69.4 \\
        TAPIS \cite{ayobi2024pixelwiserecognitionholisticsurgical} & Causal & Transformer & 73.4 \\
        Trans-SVNet \cite{gao2021transsvnetaccuratephaserecognition} & Causal & Transformer & 71.9 \\
        MuST \cite{pérez2024mustmultiscaletransformerssurgical} & Causal & Transformer & 77.3 \\
        S5 \cite{smith2023simplifiedstatespacelayers} & Causal & SSM & 71.2 \\
        Mamba \cite{gu2024mambalineartimesequencemodeling} & Causal & SSM & 71.1 \\
        Mamba-2 \cite{dao2024transformersssmsgeneralizedmodels} & Causal & SSM & 73.4 \\
        \midrule
        HID-SSM (Ours) & Causal & SSM & 80.1 \\
        HID-SSM (Ours) & Contextual & SSM & \textbf{90.2} \\
        \bottomrule
    \end{tabular}
    }
    \label{tab:heichole}
    \vspace{-4ex}
\end{table}

\subsection{Datasets and Metrics}
\noindent \textbf{Cholec80 Dataset } The Cholec80 dataset \cite{twinanda2016endonetdeeparchitecturerecognition} is a large public dataset comprising 80 videos of cholecystectomy surgeries performed by 13 surgeons. The videos were recorded at 25 fps and annotated at the same frame rate. Each frame has a resolution of either 1920×1080 or 854×480. The dataset includes seven surgical phases, we downsample the videos to 1 fps and split the dataset into 40 videos for training, 8 videos for validation, and 32 videos for testing. The final performance is reported on all 40 evaluation videos to align with the literature. 

\noindent \textbf{MICCAI2016 Dataset} The MICCAI2016 dataset \cite{twinanda2016endonetdeeparchitecturerecognition,stauder2017tumlapcholedatasetm2cai} consists of 41 laparoscopic videos of cholecystectomy procedures. The videos are all that are captured at 25fps, and each frame has a resolution of 1920 × 1080. The dataset consists of eight surgical phases. Following prior works \cite{Czempiel_2020,gao2021transsvnetaccuratephaserecognition}, we use 27 videos as training videos, 4 videos as validation videos, and 10 videos as testing videos. The final performance is reported on all 14 videos to ensure consistency with prior studies.

\noindent \textbf{Heichole Dataset} The Heichole Dataset \cite{wagner2021comparativevalidationmachinelearning} comprises 24 publicly available videos of laparoscopic cholecystectomy procedures. The dataset also contains 7 surgical phases identical to Cholec80. To ensure computational efficiency, we uniformly downsample the videos to 1 frame per second (fps). The dataset is annotated with seven surgical phases, where the phases can have complicated interleaving. We adopt the protocol established in previous work \cite{pérez2024mustmultiscaletransformerssurgical}, partitioning the dataset into 16 videos for training and 8 videos for evaluation. Specifically, within the 8 evaluation videos, 2 are used for validation, while the remaining 6 serve as the primary test set. The final performance is reported on all 8 evaluation videos to ensure consistency with prior studies.

\noindent \textbf{Metrics} For the Cholec80 and MICCAI2016 datasets, we use four evaluation metrics—accuracy (AC), precision (PR), recall (RE), and Jaccard (JA)—following previous approaches \cite{yang2024surgformersurgicaltransformerhierarchical, SKiT, liu2023lovitlongvideotransformer}. These metrics are evaluated in a relaxed manner, where a prediction is considered correct if it falls within a 10-second window of a phase transition and aligns with a neighboring phase, even if it does not exactly match the ground truth label. For the Heichole dataset, we follow the approach by \cite{pérez2024mustmultiscaletransformerssurgical} and report the unrelaxed micro-averaged F1-score.

\subsection{Implementation Details}
Our model is implemented with the PyTorch framework and runs on a single NVIDIA RTX A6000 GPU, which provides high computational precision. For all datasets, we first uniformly downsample the video to 1 fps and feed the resulting sequence of frames into the visual feature extractor, which further downsample every surgical video to \(100\) frames. As the backbone, we employ a Swin Tiny model pretrained on ImageNet-1k \cite{russakovsky2015imagenetlargescalevisual}, initializing the learning rate at \(5 \times 10^{-5}\). On top of the Swin Tiny backbone, we incorporate two layers of State Space Models (SSM) (\(N_s = 2\)), trained with a learning rate of \(2 \times 10^{-4}\). The visual feature extractor is optimized using the AdamW optimizer with hyperparameters \(\beta = \{0.9, 0.999\}\), \(\epsilon = 1 \times 10^{-8}\), and a weight decay of \(1 \times 10^{-2}\). The Feature Extractor is trained for a single epoch. Additionally, we employ a three-layer ID-SSM as PPN (\(N_q=3\), optimized using the Adam optimizer with a learning rate of \(2 \times 10^{-4}\). For the causal HID-SSM, which is constrained to only use information in the past, we set the number of \globalModels\ layers to \(N_G = 4\) for both the Cholec80 and MICCAI2016 datasets, and \(N_G = 5\) for the Heichole dataset. For the contextual HID-SSM, which explicitly leverages future information, we set the number of \globalModels\ layers to \(N_G = 5\) for the Cholec80 dataset, while using \(N_G = 4\) for both the MICCAI2016 and Heichole datasets. The number of \localModels\ layers \(N_L\) are uniformly set to \(1\). The HID-SSM is trained using the Adam optimizer with a learning rate of \(2 \times 10^{-4}\) for 20 epochs. We set the relative importance of two losses \(\alpha=0.7\) across all datasets.
\subsection{Results} 

\begin{figure*}[htbp]  
\centering  
\includegraphics[width=0.83\textwidth]{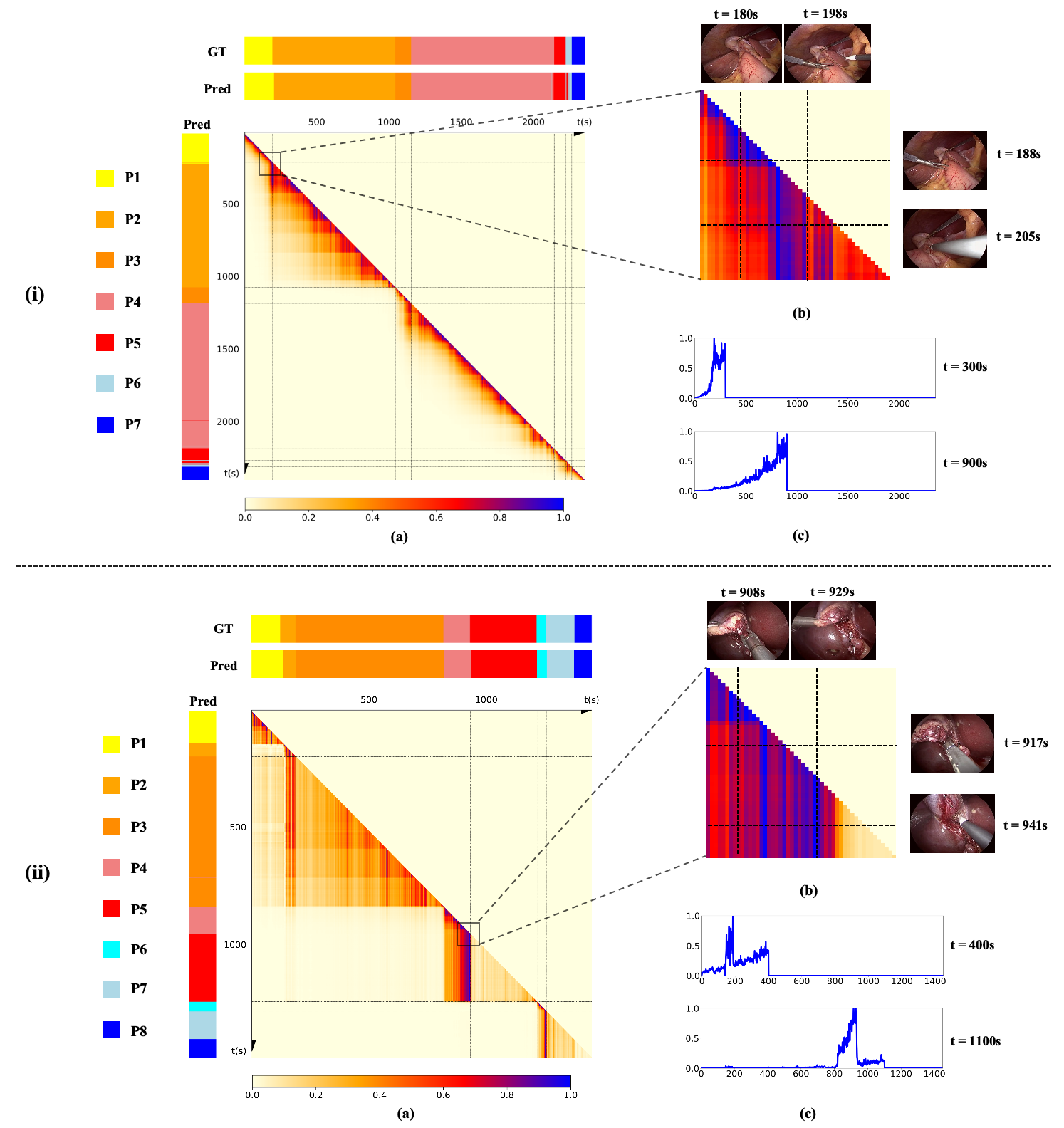}
\caption{Schematic illustration of HID-SSM’s long-term dependency modeling for two surgical videos: (i) Cholec80 Video67 and (ii) MICCAI2016 Video41. For each video: (a) Per-row normalized matrix mixer; (b) Zoom-in view of the matrix mixer; (c) Quantitative line plots of contribution weights from past frames at specified time steps \(t\).}
\label{attn_matrix}  
\vspace{-4ex}
\end{figure*} 

\begin{figure*}[htbp]  
\centering  
\includegraphics[width=\textwidth]{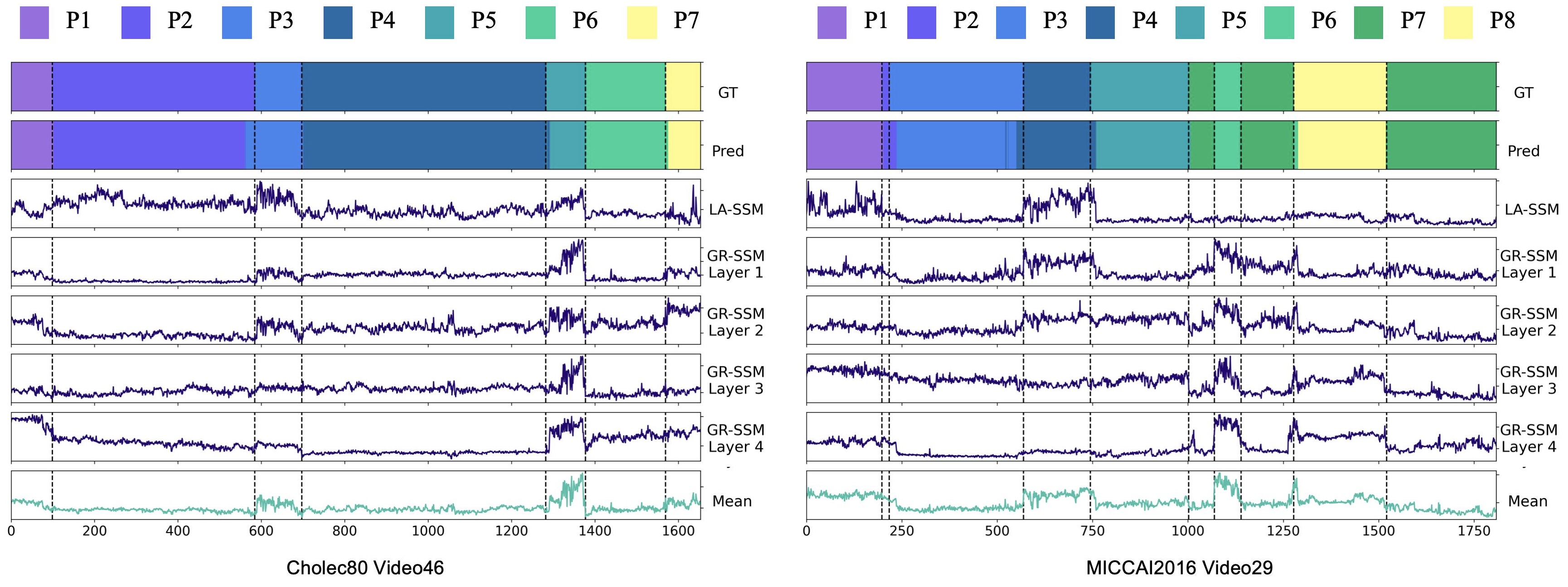} 
\caption{Visualization of $\Delta_t$ dynamics for Cholec80 Dataset Video62 (Left) and the MICCAI2016 Dataset Video41 (Right). The horizontal axis represents the timestep. The color ribbon at the top illustrates the Ground Truth (GT) and corresponding predictions (Pred). Each line chart in the subplot below visualizes the dynamics of $\Delta_t$ for different layers, with the mean $\Delta_t$ shown at the bottom. It can be observed that the $\Delta_t$ experiences a sharp change at phase transitions, as detailed in this section.}  
\label{dt_plot} 
\vspace{-4ex}
\end{figure*} 

\subsubsection{Quantitative Results}
We benchmark our model on three datasets against several state-of-the-art (SOTA) methods, including CNN-based approaches like SV-RCNet\cite{SV-RCNet} for joint spatial-temporal learning \and TMRNet\cite{jin2021temporalmemoryrelationnetwork} for long-range information retention. We also compare with Transformer-based models like Trans-SVNet\cite{gao2021transsvnetaccuratephaserecognition}, which retrieves spatial information after temporal modeling with attention; Surgformer\cite{liu2023lovitlongvideotransformer}, which attends to local relationships between the current frame and various temporal resolutions of past frames. Finally, we evaluate SSM-based baselines such as S5\cite{smith2023simplifiedstatespacelayers}, Mamba\cite{gu2024mambalineartimesequencemodeling}, and Mamba-2\cite{dao2024transformersssmsgeneralizedmodels}.

We first quantitatively evaluate our HID-SSM on Cholec80 dataset. As shown in Table \ref{tab:cholec80}, our causal HID-SSM—constrained to use only past and present information—delivers competitive performance, achieving a mean accuracy of 94.5\% and a mean precision of 92.1\%, surpassing other causal models.While our model’s recall and Jaccard scores are slightly lower than those of Surgformer \cite{yang2024surgformersurgicaltransformerhierarchical}, with mean values of 91.1\% and 83.8\%, respectively, our contextual HID-SSM, which explicitly leverages future information, significantly outperforms all previous methods across all evaluation metrics. It achieves a mean accuracy of 96.2\%, mean precision of 93.3\%, mean recall of 93.1\%, and mean Jaccard of 86.5\%, demonstrating superior performance in all evaluation metrics. As shown in Table \ref{tab:Cholec80phase}, the model performs generally well across all phases, demonstrating the best performance during the two longest phases: Calot Triangle Dissection and Gallbladder Dissection. This highlights the HID-SSM's strong capacity to retain long-term information. Additionally, the consistent performance across these phases further supports the model's ability to capture in-phase dynamics, thereby confirming its proficiency in modeling local information.

On the MICCAI2016 dataset, we adopt a similar approach to benchmark our method. As shown in Table \ref{tab:m2cai}, our causal HID-SSM outperforms the current state-of-the-art methods across all metrics, achieving a mean accuracy of 91.5\%, mean precision of 90.2\%, mean recall of 89.8\%, and mean Jaccard of 80.2\%. Additionally, our contextual HID-SSM achieves slightly superior results, with a mean accuracy of 91.1\%, mean precision of 90.7\%, mean recall of 89.1\%, and mean Jaccard of 81.7\%. As shown in Table \ref{tab:MICCAIphase}, our model performs consistently well across all 8 phases, with particularly strong results in the Calot Triangle Dissection and Gallbladder Dissection phases. Notably, our HID-SSM effectively captures complex phase interleaving, especially between the last three phases: P6 (Gallbladder Packaging), P7 (Cleaning Coagulation), and P8 (Gallbladder Retraction) as detailed in Fig.\ref{dt_plot}. This demonstrates our model’s effectiveness in balancing global and local information. It learns the general relationships between phase appearances while also capturing the fine-grained details that signify phase transitions, allowing the model to make holistic accurate predictions.

We further evaluate our method's F1-score on the Heichole dataset. As shown in Table \ref{tab:heichole}, our causal HID-SSM outperforms the state-of-the-art methods, achieving an F1 score of 80.1\%. Additionally, our contextual HID-SSM demonstrates a substantial performance boost, achieving an F1 score of 90.2\%. Given that the Heichole dataset contains only 16 training videos—far fewer than Cholec80 and MICCAI2016—and even more complex phase interleaving, our superior results on Heichole further validate the model's ability to capture intricate phase transitions with less data.

\subsubsection{Qualitative Results}
We further justify the effectiveness of HID-SSM in surgical phase recognition by visualizing its ability to handle phase jumps and model long-range dependencies. Specifically, we highlight how HID-SSM effectively manages sudden transitions between phases and captures long-range dependencies, demonstrating its robustness in handling complex, dynamic surgical scenarios.
\\
\textbf{Long-Term Dependency Awareness}: As discussed in \S \ref{HID-SSM}, each \localModels\ and \globalModels\ layer can be represented in the matrix mixer form \eqref{eq9:GR-SSM}\eqref{eq10:LA-SSM}, where the output \( y_i \) at frame \( i \) is determined by the multiplication of the \( i^{th} \) row of the matrix mixer \( \mathbf{M} \) with the entire input \( u \). Thus, each row of \( \mathbf{M} \) can be interpreted as an attention assigned to the corresponding input at frame \( i \). Building on this observation, we visualize the matrix mixer \( \mathbf{M} \) of the last layer of GR-SSM, and apply min-max normalization to each row, as illustrated in Fig.~\ref{attn_matrix}(a).

In Fig.~\ref{attn_matrix}, panel (a) visualizes the matrix mixer alongside the phase map, this demonstrates our model's capabilities to capture long dependencies. Panel (b) zooms in on the matrix mixer, which showcases the regions with the highest attention scores. Additionally, to highlight how the matrix mixer represents the contribution of each surgical frame to the current prediction, we plot the contribution rate of past frames for selected timesteps in panel (c), which is obtained by simply taking the corresponding row of the matrix mixer. Specifically, the contribution rate map is generated by extracting the corresponding row from the matrix mixer and visualizing its values as a line chart. As can be observed from Fig.~\ref{attn_matrix}, on both Cholec80 (i)(a) and MICCAI2016 (ii)(a), the model effectively captures long-range temporal dependencies by retrieving information from frames occurring several hundred timesteps earlier when making the decision for the current frame. This is justified by (i)(c) and (ii)(c), which exemplified how the current prediction is affected by inputs hundred of timesteps before. For example, as illustrated in (i)(c), when predicting the phase at timestep \( t = 300 \), the model assigns a high attention weight to timestep \( t = 198 \) (second frame from the top left in (i)(b)). Notably, this frame marks the first appearance of a new hook tool in the video, indicating that the model recognizes and prioritizes key changes. This finding underscores the model's ability to attend to salient temporal features over extended sequences, thereby demonstrating its efficacy in capturing long-term dependencies within surgical workflow analysis.
\\
\textbf{Temporal Selective Activation}: As revealed in \eqref{eq:dt}, when $\Delta_t$ is large, the HID-SSM resets its state and prioritizes the current surgical input \( u_t \), while a small $\Delta_t$ preserves the state and disregards \( u_t \). To analyze this effect, we first demonstrate the ground truth and predicted phase map using color ribbons. We then visualize the $\Delta_t$ values across HID-SSM layers in Fig.~\ref{dt_plot}. Specifically, for each \localModels\ layer, we apply Z-score normalization to each \(\mathcal{SS}\) block and concatenate the $\Delta_t$ values from all \( K \) blocks to obtain the overall $\Delta_t$ dynamics. For \globalModels\, we simply visualize the $\Delta_t$ values for each \globalModels\ layer without any normalization. We also visualize the mean value of all \localModels\ and \globalModels\ layers by taking the average of those layers.

From Fig.~\ref{dt_plot}, it is evident that near phase transitions, the $\Delta_t$ values for all layers exhibit a sharp decline, signaling that the HID-SSM temporarily disregards the current input in anticipation of a phase jump. Following this decline, a sharp increase in $\Delta_t$ occurs, indicating a reset of the previous state, which facilitates adaptation to the new phase. Notably, this sharp variation in $\Delta_t$ aligns well with the phase transition period, underscoring its ability to effectively capture abrupt transitions in the temporal dynamics. 
%In addition, \localModels\ exhibit more complex $\Delta_t$ dynamics within phases, reflecting their enhanced capacity to selectively attend to relevant temporal information. This selective attention enables the \localModels\ to handle intricate local temporal dependencies more effectively, thus facilitating the comprehensive modeling of \globalModels.

\section{Limitation}
The primary limitation of our proposed method arises from the inherent non-deterministic behavior of some ID-SSM \cite{dao2024transformersssmsgeneralizedmodels, hwang2024hydrabidirectionalstatespace}, which currently lack a comprehensive solution\footnote{https://github.com/state-spaces/mamba/issues/137}. Specifically, these ID-SSMs implement atomic adds, which introduce non-determinism into the backward pass.

In our experiments, we explored various strategies to mitigate the effects of this randomness, including reducing the learning rate, training for fewer epochs, and utilizing higher precision floating-point arithmetic. These measures were effective in minimizing the randomness in HID-SSM, with performance fluctuations generally confined to around 0.1\%. However, we have yet to identify a robust method to fully control the randomness in our visual feature extractor , where randomness in the ID-SSM head propagates deeply through the Swin-Transformer, leading to performance fluctuations in the final model at around 1\%.

\section{Conclusion}
In this paper, we propose a novel HID-SSM framework for surgical workflow analysis. Unlike previous methods that rely on limited temporal context, our model processes the entire surgical video to produce full-length phase predictions. Within HID-SSM, we introduced three key components: a temporally consistent visual feature extractor which integrates global temporal information early in the feature extraction; A \localModels\ module, which captures fine-grained local temporal patterns, and a \globalModels\ module, which holistically processes the spatial-temporal enriched representation. Together, these blocks construct a hierarchically structured video representation that effectively encodes both local and global dependencies. Extensive experiments on three benchmark datasets demonstrate that HID-SSM outperforms existing state-of-the-art methods, both quantitatively and qualitatively, showing superior capabilities in handling dynamic phase transitions and maintaining awareness of long-range dependencies, confirming its effectiveness for surgical video understanding.

\bibliographystyle{ieeetr}
\small\bibliography{tmi}
\onecolumn
\appendix

\begin{figure*}[htbp]
\centering  
\includegraphics[width=0.8\textwidth]{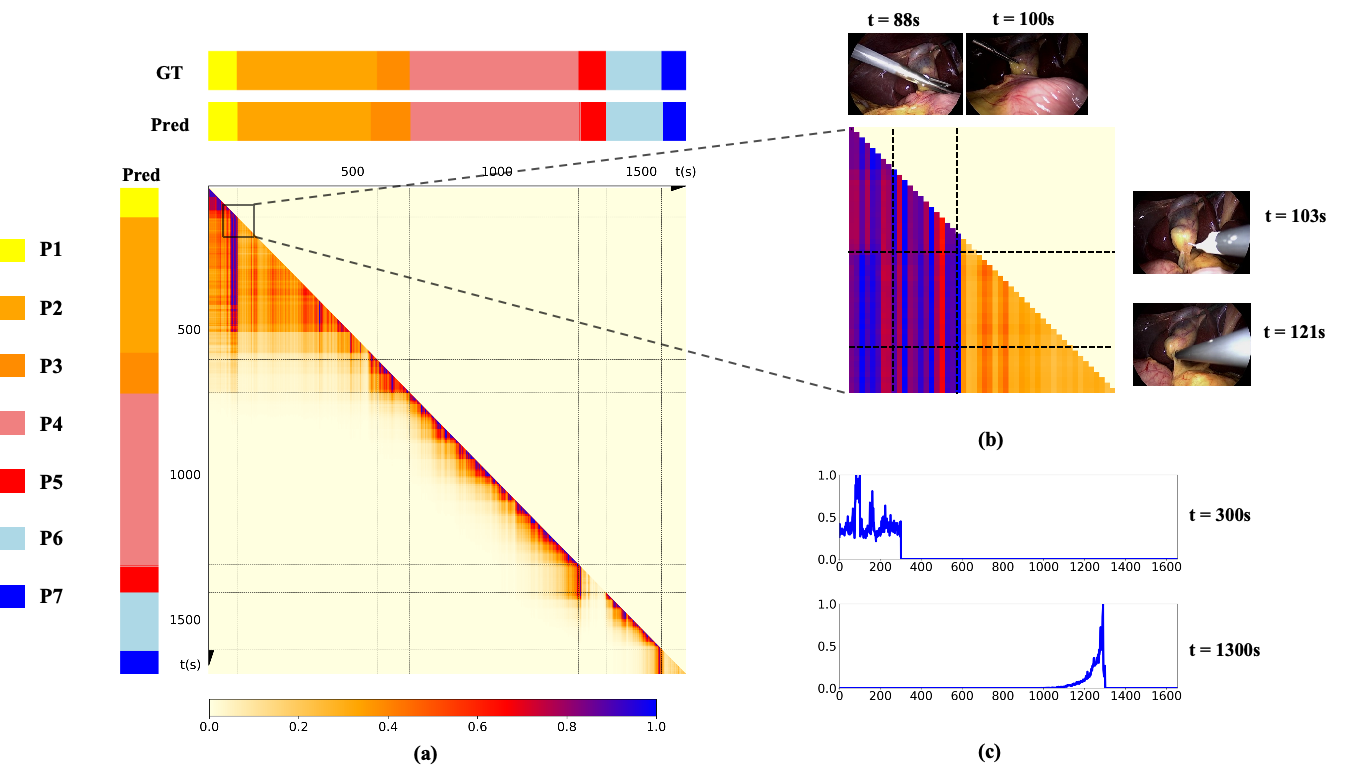} 
\caption{Schematic illustration of HID-SSM’s long-term dependency modeling for Cholec80 Video46; (a) Per-row normalized matrix mixer; (b) Zoom-in view of the matrix mixer; (c) Quantitative line plots of contribution weights from past frames at specified time steps \(t\) }  
\label{attn_matrix_cholec52}  
\end{figure*}  

\begin{figure*}[htbp]  
\centering  
\includegraphics[width=0.8\textwidth]{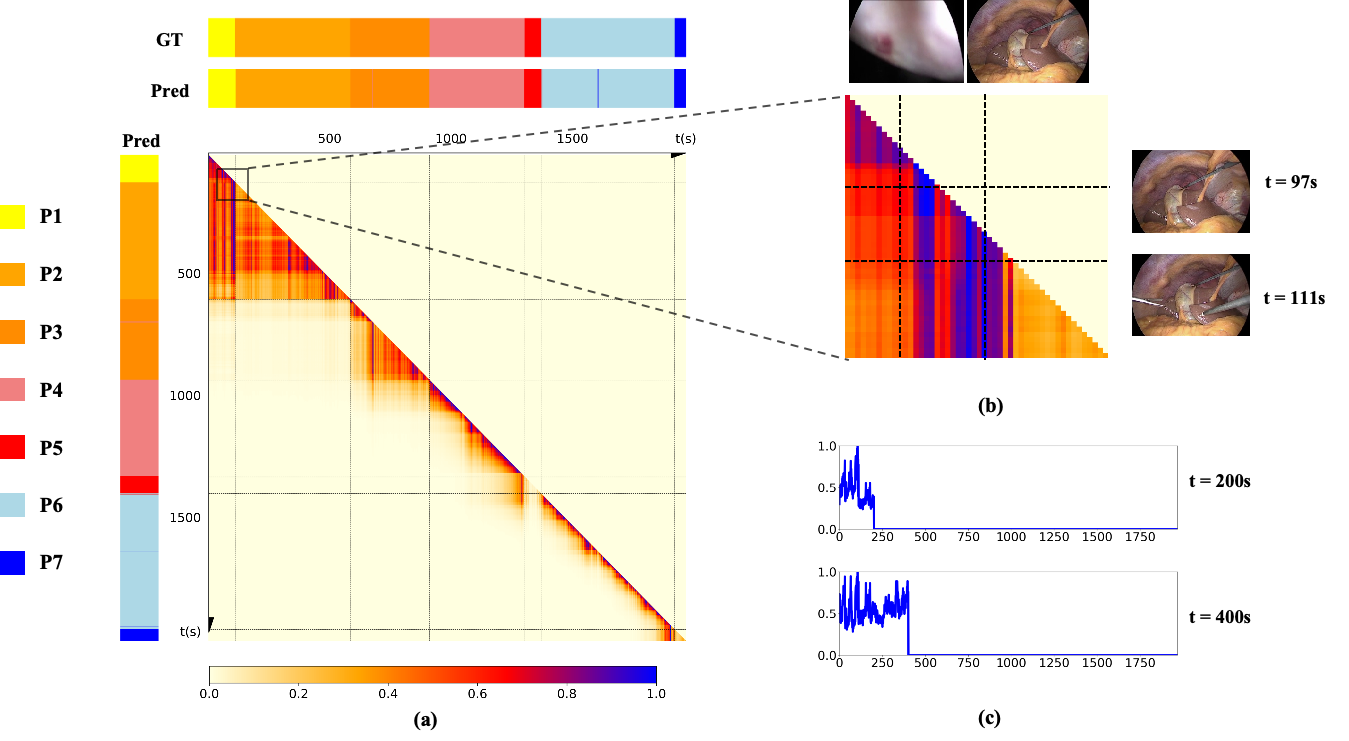} 
\caption{Schematic illustration of HID-SSM’s long-term dependency modeling for Cholec80 Video52; (a) Per-row normalized matrix mixer; (b) Zoom-in view of the matrix mixer; (c) Quantitative line plots of contribution weights from past frames at specified time steps \(t\) }  
\label{attn_matrix_cholec67}  
\end{figure*} 

\begin{figure*}[htbp]  
\centering  
\includegraphics[width=0.8\textwidth]{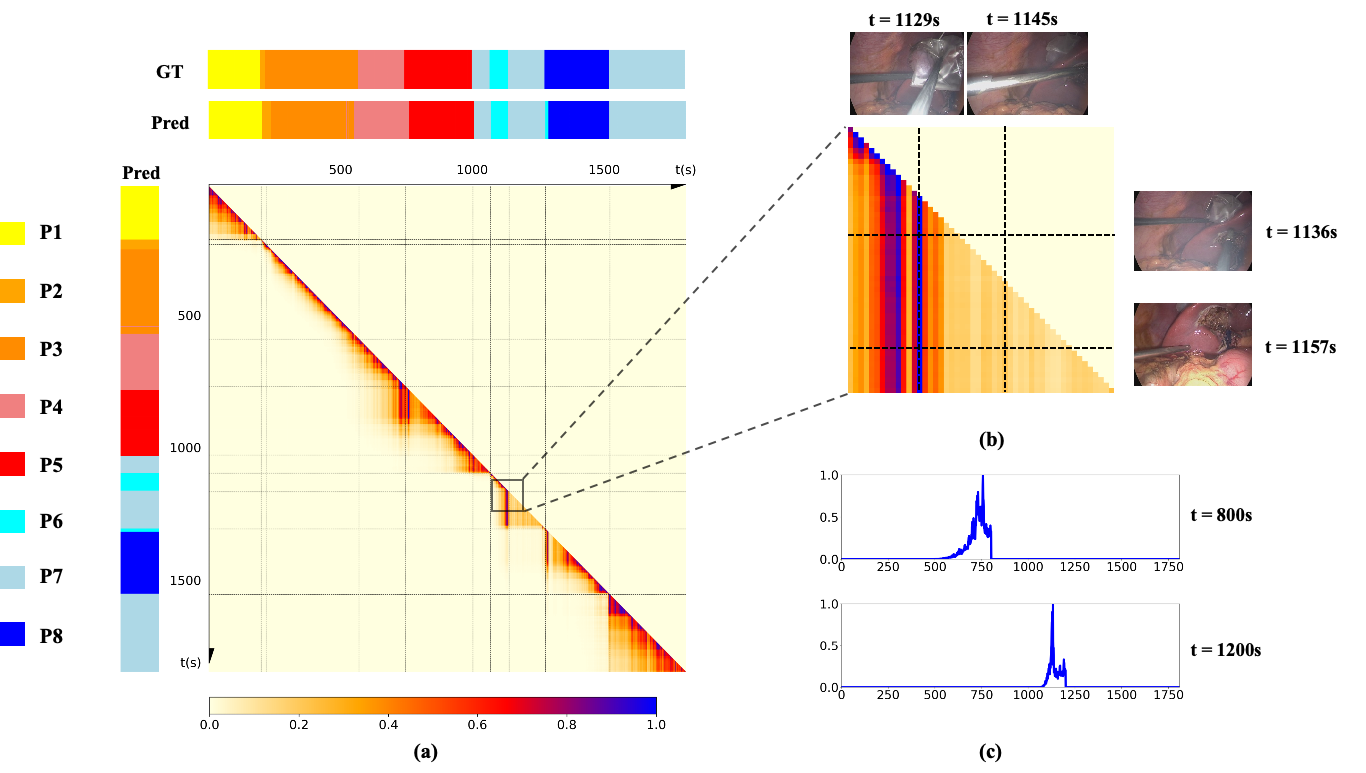} 
\caption{Schematic illustration of HID-SSM’s long-term dependency modeling for MICCAI2016 Video29; (a) Per-row normalized matrix mixer; (b) Zoom-in view of the matrix mixer; (c) Quantitative line plots of contribution weights from past frames at specified time steps \(t\) }  
\label{attn_matrix_miccai29}  
\end{figure*} 

\begin{figure*}[htbp]  
\centering  
\includegraphics[width=0.8\textwidth]{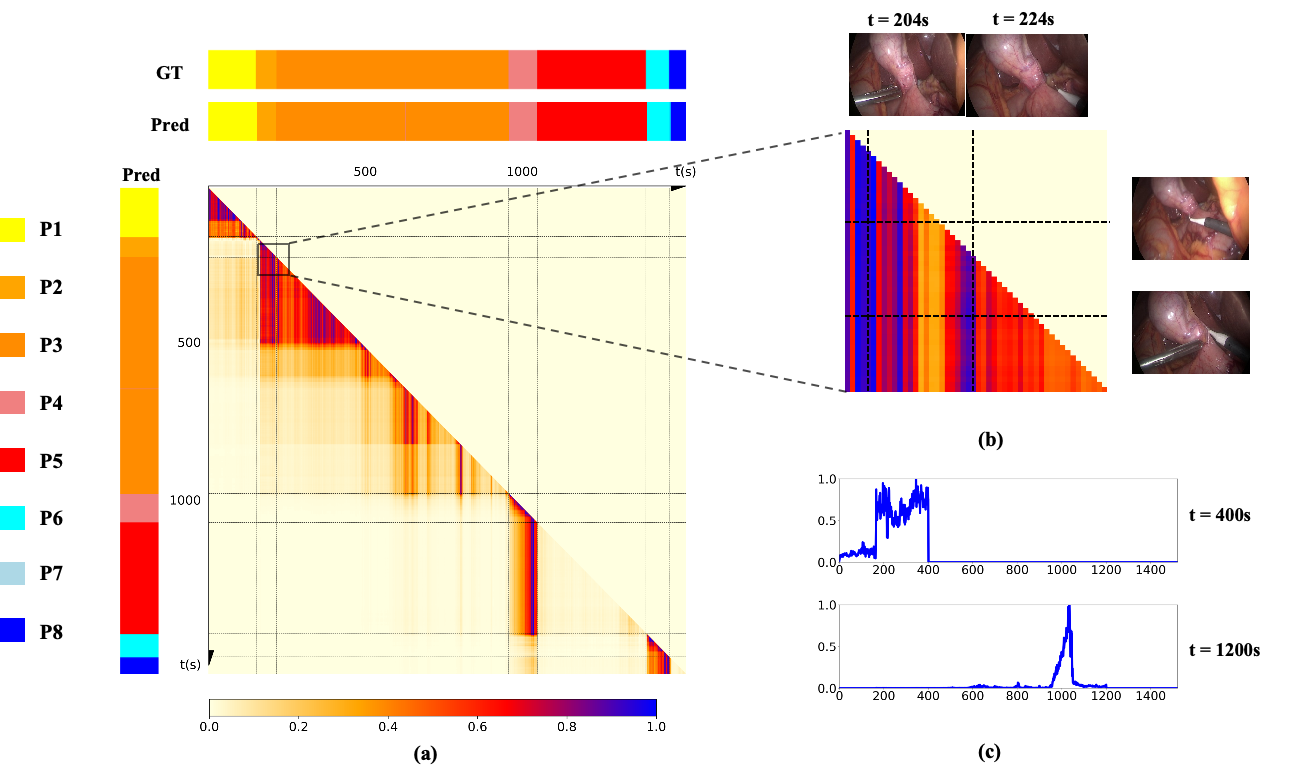} 
\caption{Schematic illustration of HID-SSM’s long-term dependency modeling for MICCAI2016 Video34; (a) Per-row normalized matrix mixer; (b) Zoom-in view of the matrix mixer; (c) Quantitative line plots of contribution weights from past frames at specified time steps \(t\) }  
\label{attn_matrix_miccai34}  
\end{figure*} 
\clearpage
\twocolumn
\begin{table}[ht]
\centering
\caption{Per‐phase performance metrics for Causal HID-SSM on Cholec80 Dataset}
\label{tab:Cholec80phase}
\begin{tabular}{@{}lrrr@{}}
\toprule
Phase                         & Jaccard (\%) & Precision (\%) & Recall (\%) \\ 
\midrule
Preparation                   & 88.34     & 97.96     & 90.39     \\
Calot Triangle Dissection       & 95.74     & 97.89     & 97.57     \\
Clipping Cutting               & 85.46     & 90.94     & 93.55     \\
Gallbladder Dissection         & 92.28     & 94.60     & 98.15     \\
Gallbladder Packaging          & 80.74     & 90.03     & 90.23     \\
Cleaning Coagulation           & 65.87     & 89.04     & 73.35     \\
Gallbladder Retraction         & 77.99     & 84.32     & 94.13     \\
\bottomrule
\end{tabular}
\end{table}
\newpage
\begin{table}[ht]
\centering
\caption{Per‐phase performance metrics for Causal HID-SSM on MICCAI2016 Dataset}
\label{tab:MICCAIphase}
\begin{tabular}{@{}lrrr@{}}
\toprule
Phase                         & Jaccard (\%) & Precision (\%) & Recall (\%) \\ 
\midrule
Trocar Placement               & 85.97     & 94.69     & 91.02     \\
Preparation                   & 70.20     & 84.24     & 89.30     \\
Calot Triangle Dissection       & 96.88     & 98.98     & 97.94     \\
Clipping Cutting               & 83.90     & 94.12     & 90.58     \\
Gallbladder Dissection         & 89.12     & 92.33     & 98.12     \\
Gallbladder Packaging          & 68.59     & 90.97     & 76.40     \\
Cleaning Coagulation           & 71.29     & 81.87     & 84.23     \\
Gallbladder Retraction         & 75.84     & 84.73     & 90.95     \\
\bottomrule
\end{tabular}
\end{table}

\end{document}